\documentclass[10pt,twocolumn,letterpaper]{article}

\usepackage{iccv}              
\usepackage{soul}   
\usepackage[accsupp]{axessibility}
\definecolor{iccvblue}{rgb}{0.21,0.49,0.74}
\usepackage[pagebackref,breaklinks,colorlinks,allcolors=iccvblue]{hyperref}

\def\paperID{13997} 
\def\confName{ICCV}
\def\confYear{2025}

\title{Semi-supervised Deep Transfer for Regression without Domain Alignment}

\author{
Mainak Biswas$^{1, 2}$, Ambedkar Dukkipati$^{3}$,  Devarajan Sridharan$^{1, 2, 3}$ \\
$^{1}$Brain, Computation and Data Sciences, $^{2}$Centre for Neuroscience, $^{3}$Computer Science and Automation \\
Indian Institute of Science, Bangalore - 560012 \\
{\tt\small mainakbiswas@iisc.ac.in, ambedkar@iisc.ac.in, sridhar@iisc.ac.in}
}
\newcommand{\ud}{\mathrm{d}}
\begin{document}
\maketitle
\begin{abstract}
Deep learning models deployed in real-world applications (e.g., medicine) face challenges because source models do not generalize well to domain-shifted target data. Many successful domain adaptation (DA) approaches require full access to source data or reliably labeled target data. Yet, such requirements are unrealistic in scenarios where source data cannot be shared either because of privacy concerns or because it is too large and incurs prohibitive storage or computational costs. Moreover, resource constraints may limit the availability of labeled targets. We illustrate this challenge in a neuroscience setting where source data are unavailable, labeled target data are meager, and predictions involve continuous-valued outputs. We build upon Contradistinguisher (CUDA), an efficient framework that learns a shared model across the labeled source and unlabeled target samples, without intermediate representation alignment. Yet, CUDA was designed for unsupervised DA, with full access to source data, and for classification tasks. We develop CRAFT -- a Contradistinguisher-based Regularization Approach for Flexible Training -- for source-free (SF), semi-supervised transfer of pretrained models in regression tasks.  We showcase the efficacy of CRAFT in two neuroscience settings: gaze prediction with electroencephalography (EEG) data and ``brain age'' prediction with structural MRI data. For both datasets, CRAFT yielded up to $9\%$ improvement in root-mean-squared error (RMSE) over fine-tuned models when labeled training examples were scarce. Moreover, CRAFT leveraged unlabeled target data and outperformed four competing state-of-the-art source-free domain adaptation models by more than $3\%$. Lastly, we demonstrate the efficacy of CRAFT on two other real-world regression benchmarks. We propose CRAFT as an efficient approach for source-free, semi-supervised deep transfer for regression that is ubiquitous in biology and medicine.



\end{abstract}

\hspace{-4mm}{\bf Keywords: } source-free DA, continuous label prediction, gaze prediction, brain age prediction, healthcare    
\section{Introduction}
\label{section:Introduction}
For the successful application of deep learning models in the real world, they must be robust to ``domain shift''~\cite{domain_shift_problems}. For example, in biology and medicine, supervised deep learning models are often trained on large, high-quality source datasets collected in resource-rich settings~\cite{Miller2016, Eyepacs}. However, naive transfer or even simple finetuning fails to produce high accuracies when tested with smaller target datasets of mixed quality collected~\cite{Sundarakumar2022, aptos}. As a result, state-of-the-art domain adaptation (DA) approaches, which enable the source models to generalize successfully to target datasets, must be employed. Many of these approaches involve explicit alignment of source and target, while learning a shared predictive model for both domains~\cite{Ganin2016, Park2020, cmd_iclr}. 

A particular challenge occurs with DA when the source dataset is unavailable. This can happen either because of privacy or proprietary concerns, such as with medical images or patents \cite{privacy}. Alternatively, source datasets may be too large, incurring prohibitive storage and computation costs in low-resource settings~\cite{Miller2016}. Source-free (SF) domain adaptation methods that address these challenges are getting increasingly popular~\cite{review_sf1, review_sf2}. Such methods generalize the source model to unlabeled or partially target data (unsupervised or semi-supervised DA) with one of two approaches: data-based or model-based. While data-based methods create surrogate source data, model-based methods adapt the source model to the target domain (see Section~\ref{section: related works}).

\begin{figure}[t]
\begin{center}
\includegraphics[scale=0.46]{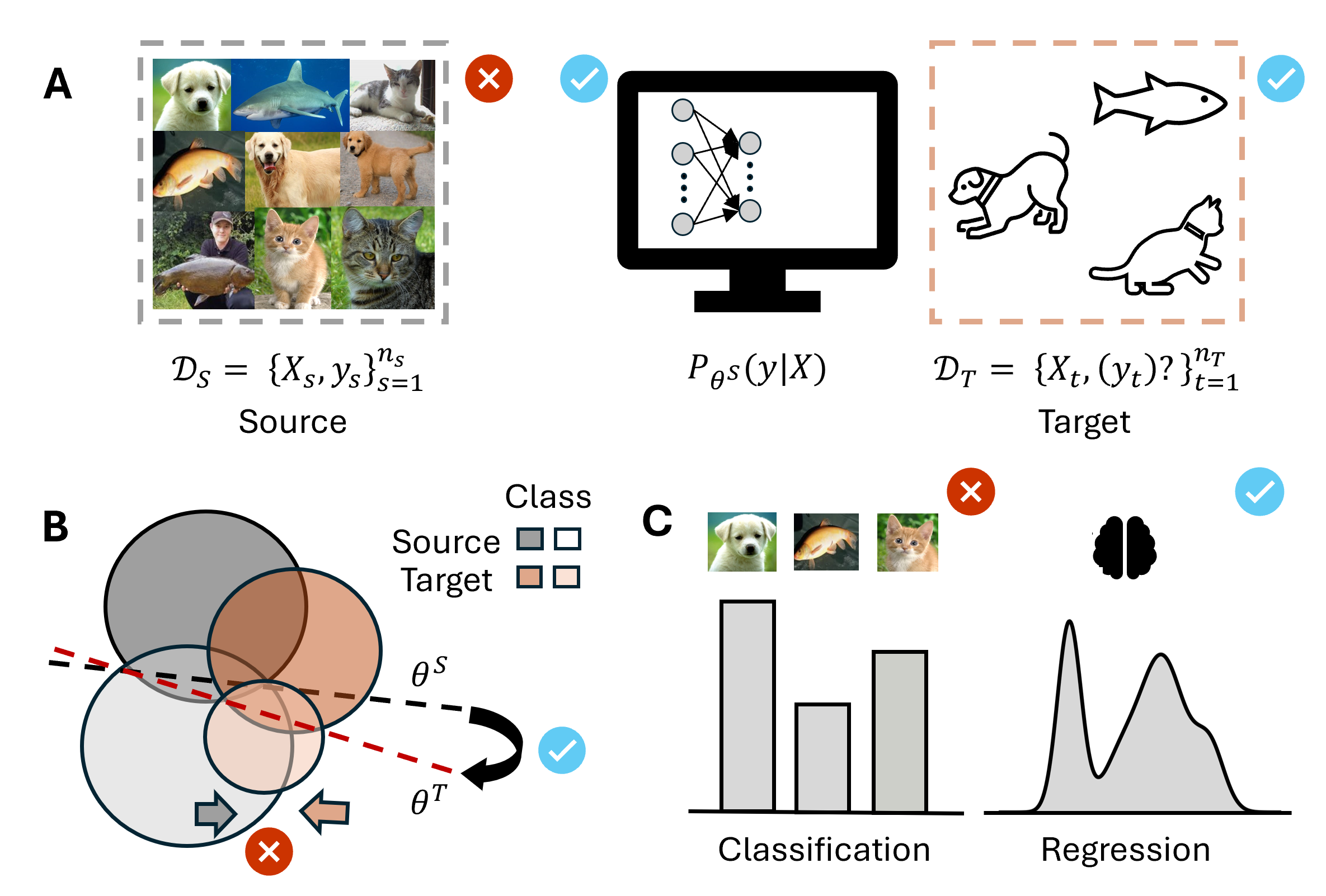}
\end{center}
\caption{Schematic of Semisupervised Deep Transfer without Domain Alignment. ({\bf A}) The approach seeks to transfer a pre-trained model to a target domain with limited labeled data when the source dataset is unavailable. ({\bf B}) Because the source dataset is unavailable, we tune the decision boundaries without explicitly aligning the representation of the domains, combining a supervised loss and the CUDA loss (regularizer). ({\bf C}) The algorithm targets deep transfer in regression tasks, like brain age prediction. }
\label{figure:introduction}
\end{figure}

We address model-based SF-DA in a semi-supervised setting. Specifically, we focus on neuroscience applications where source data are inaccessible, and target labels are reliably available only for a few samples. Regression tasks -- like predicting brain age (Fig.~\ref{figure:introduction}C)~\cite{Miller2016}, saccades~\cite{Kastrati2021}, neural activity~\cite{gifford2025algonautsproject2025challenge}, and stimulus orientation~\cite{orientationDecoding} -- are ubiquitous in neuroscience; we tackle a subset of these. We build upon Contradistinguisher~\cite{2019:BalgiDukkipati:2019:Cuda,Balgi2019}, a recently-proposed tool for effective unsupervised DA (CUDA). Unlike popular representation alignment DA methods~\cite{Ganin2016, Park2020}, CUDA directly learns a common model for both domains. Yet, CUDA requires full access to source data and addresses classification tasks. Therefore, it is not directly applicable to our problem.

Here, we leverage CUDA to develop a new algorithm --  Contradistinguisher-based Regularization Approach for Flexible Training (CRAFT) which performs deep transfer, without domain alignment, even when the source dataset is unavailable (Fig.~\ref{figure:introduction}A-B). We also extend CUDA to a regression setting. Previous studies approach regression tasks naively by binning continuous labels into discrete bins~\cite{Park2020}, sometimes employing a rank-ordinal objective~\cite{Cao2020}. Regardless, the efficacy of such approaches may depend critically on bin sizes. Our CRAFT model uses a principled approach to extend CUDA to predictions involving continuous-valued outputs. We employ binning as a practical strategy to generate and optimize pseudo-labels, without constraining the model to produce discrete outputs. Our study makes the following main contributions.

\begin{itemize}
\item First, we extend CUDA to source-free DA for regression tasks involving continuous-valued outputs. 
\item Second, we formulate a theoretically-motivated semi-supervised objective by combining supervised and unsupervised losses with a tunable hyperparameter ($\alpha$) (Section~\ref{section:The CRAFT Objective}). We also derive the unsupervised loss as a maximum entropy prior to model parameters (Appendix~\ref{appendix: The Prior Distribution}). 
\item Third, we show the effectiveness of CRAFT on two neuroscience challenges -- gaze prediction from electroencephalographic (EEG)~\cite{Kastrati2021} data and brain age prediction from magnetic resonance imaging (MRI) data~\cite{Sundarakumar2022}.
\item Fourth, we also apply CRAFT to two real-world regression benchmarks involving people counting~\cite{JHU, gao2020nwpu} and tumor size prediction~\cite{cam16, cam17}. 
\item Fifth, we compare CRAFT against state-of-the-art SF-DA methods~\cite{TASFAR, Datafree, progressiveMixup, BBCN} and show that the best improvements occur as unlabeled target samples increases, demonstrating its efficacy for semi-supervised SF-DA.
\item  Finally, we analyze the computational complexity and demonstrate numbers competitive with sota models.
\item Our code is publicly available (see Appendix~\ref{appendix: code}).
\end{itemize}

\section{Related Work}
\label{section: related works}
{\bf Domain alignment methods.} Popular Unsupervised Domain Adaptation (UDA) approaches rely on learning common intermediate representations across source and target domains. Domain Adversarial training of Neural Networks (DANN)~\cite{Ganin2016} uses a discriminator to learn common representations across both domains adversarially. In Importance Weighting~\cite{Park2020} (IW), higher importance is afforded to losses corresponding to source samples more likely to belong to the target domain. Central Moment Discrepancy (CMD)~\cite{cmd_iclr} minimizes the central moments of the latent representations between the domains, while others optimize their Maximum Mean Discrepancy (MMD)~\cite{joint_feat, mmd}. Despite being effective, they cannot be used when the source is unavailable and will not be discussed further.

\noindent{\bf Source-free UDA.} To address DA scenarios where source data and its target labels are unavailable, several source-free unsupervised DA approaches have been recently proposed. Broadly, they are categorized as: Data-based and Model-based~\cite{review_sf1}. Data-based methods generate a source-like distribution ~\cite{surrogateData, kurmi2021domainimpressionsourcedata}. However, they often depend on the availability of generative models trained on the source, which could be unavailable. For example, SF-UDA through domain‐specific perturbation (AUGFree)~\cite{augfree} assumes the existence of domain-invariant features for prediction. Since the source is unavailable, AUGFree perturbs the target data and then learns a transformation to the invariant features by adversarially aligning the target and the perturbed target.

On the other hand, model-based methods finetune the source model on the target using strategies like contrastive learning~\cite{contrastive_sfssuda}, pseudo-labeling~\cite{TASFAR}, or entropy minimization~\cite{entropyMinm}. They are often prone to overfitting to pseudo-labels. For example, Class-Balanced Multicentric Dynamic Prototype Strategy for SF-UDA (BMD)~\cite{bmd} follows a pseudo-labeling strategy using class prototypes based on the source model predictions, and in turn uses these labels for training. Source Data-Absent UDA Through Hypothesis Transfer (SHOT++)~\cite{shot} combines three objectives -- a) maximizing mutual information between the predictions and inputs, b) self-supervised loss based on prototype-based clusters, and c) a Mixup-based (See SF-SSDA) loss~\cite{progressiveMixup}. Attracting and Dispersing (AaD)~\cite{aad} defines a neighboring cluster (K nearest neighbors) and a randomly selected background cluster for each sample, and maximizes the log-likelihood of the ratio of their distributions. 

Target-agnostic SF UDA for Regression Tasks (TASFAR)~\cite{TASFAR} works on the principle that prediction uncertainty is correlated with prediction error. It uses Monte-Carlo dropout to estimate label uncertainty and divides the target dataset into confident and uncertain samples. A discrete density map is estimated using the confident data, which is then used as weights to recalibrate the pseudo-labels of uncertain data. SF-UDA to measurement Shift via Bottom-Up Feature Restoration (BUFR/DataFree)~\cite{Datafree} is designed on the principle of intermediate feature alignment of the source and the target distributions. During adaptation, the model minimizes the symmetric KL divergence between feature distributions of source (fixed) and target data.

Here, we compare our model's performance against two of the above methods - TASFAR and DataFree. TASFAR was selected as it is a pseudo-labeling-based method suitable for regression. On the other hand, DataFree is a pseudo-label-free feature alignment method, not constrained to classification tasks. By adding a supervised loss (Section~\ref{subsection: Competing Methods}), we evaluate these as semi-supervised models.

\noindent{\bf Source-free SSDA.} 
Source-free semi-supervised domain adaptation (SF-SSDA) models best address the scenario of SF-DA when few target labels are available. For example, Generalized SF-SSDA~\cite{gensssfuda} minimizes source loss alongside an unsupervised target loss which maximizes the cosine similarity between predictions of k-nearest neighbors. The model is then transferred by adversarial training based on source features generated using a Conditional VAE. Similarly,~\cite{Hao2021SourcefreeUD} proposes generating surrogate source data from target samples using gradient-based optimization. Learning from Different Samples for SF-SSDA~\cite{huang2024learningdifferentsamplessourcefree} minimizes the contrastive loss on the predicted probabilities of the target domain. Furthermore, it minimizes supervised and contrastive loss on a set comprising labeled and unlabeled data with low uncertainty. Likewise, Mutual Enhancement training for SS Hypothesis transfer (MESH)~\cite{entropyMinm} minimizes the supervised loss on the labeled samples and entropy on pseudo-labeled samples. Finally, the target manifold is smoothened by making predictions robust to perturbations.

Progressive Mixup~\cite{progressiveMixup} is trained on augmentations by convex combinations of data on which the source model is confident and ones on which it is uncertain (hybrid mixup). Similarly, another set is obtained by convex combinations of samples in the (pseudo) labeled dataset (self-mixup). In contrast, Bilateral-Branch Consistency Network (BBCN)~\cite{BBCN} computes prototypes as the weighted mean of feature representations of the target dataset by the source model. Pseudo-labels are selected based on the proximity of unlabeled data to these prototypes. A separate target branch is trained using the pseudo-labels, and a consistency loss is added between the representations of both branches. The need for class-level prototype generation makes it challenging to adapt this model to regression tasks.

We compare the performance of CRAFT with two SF-SSDA methods -- SF-SSDA via progressive Mixup~\cite{progressiveMixup}, and BBCN~\cite{BBCN}, as they have little dependence on the source distribution and its derivatives. Moreover, these methods are based on complementary ideas, and the latter~\cite{BBCN} has been applied specifically in medical imaging, for tuberculosis recognition. 
\section{The CRAFT Model}
\label{section:The CRAFT Objective}
\subsection{Model formulation}
{\bf Background}. Our objective is to develop an approach for source-free semi-supervised domain adaptation, for regression tasks. For this, we build upon the {\it Contradistinguisher-based Unsupervised Domain Adaptation (CUDA)} model \cite{Balgi2019}. CUDA obviates the need for learning a common intermediate representation across the source and the target datasets. It optimizes a joint distribution over the target features and category labels to adapt the source model. Because CUDA is intended for unsupervised domain adaptation (UDA), target labels are unavailable. Hence, the algorithm uses a two-step joint optimization over the target labels and model parameters. First, it generates pseudo-labels on the target dataset from the source model~\cite{Grandvalet2004}. Then, it updates the model parameters by fixing the pseudo-labels. 
 
In this study, we extend CUDA to a source-free semi-supervised setting (SF-SSDA). Briefly, to the supervised label loss, we add a `regularized'', unsupervised loss as a prior on the model parameters.  
As we will show in the next section, the model can be flexibly applied, either in SF-UDA case or in the SF-SSDA case, depending on the weightage given to the supervised loss. We, moreover, extend the model to continuous label prediction. We call our new formulation CRAFT - {\bf C}ontradistinguisher-based {\bf R}egularization {\bf A}pproach for {\bf F}lexible {\bf T}raining (CRAFT). 

\noindent {\bf Objective. }
We consider a task in which a source model $\theta^s$ trained on a source dataset -- involving continuous label prediction (regression) -- is already available to us. The target dataset 
comprises i.i.d samples from an unknown distribution $p^t ({\bf x}, y)$: $\mathcal{D}^t = \{{\bf x}_i^t, y_i^t\}_{i=1}^{N_l}, \{{\bf x}_i^t\}_{i=1}^{N_{ul}}$ where ${{\bf x}_i^t \in \mathbb{R}^d}$ is a feature-vector, and ${y_i^t \in \mathbb{R}}$ is its corresponding label. Target training data are partially and sparingly labeled; we define the proportion of unlabeled samples ($N_{ul}$) to the total number of samples ($N$), as $N_{ul}:(N_l + N_{ul})$.

We formulate, first, the supervised objective. For continuous label prediction problems, the conditional density of the label given the features is assumed to be normally distributed. The mean is modeled using a neural network $f(. ; \theta)$, parameterized by $\theta$, as: $p (y^t | {\bf x}^t, \theta) = \mathcal{N} (y^t; f({\bf x}^t; \theta), c)$, and the variance, $c$ is assumed to be constant as it simplifies the optimization. In the final formulation, the variance occurs as the weight of the unsupervised loss, $\alpha$ (equation~\ref{equn:craft-objective}); therefore, only one of the two hyperparameters ($c$ or $\alpha$) needs to be tuned. Hence, the log-likelihood can be written as:

\begin{equation}
\text{log } p (\mathcal{D}^t | \theta) \propto - \sum_{i=1}^{N_l}  (y_i^t - f({\bf x}_i^t; \theta))^2  
\label{equn:likelihood}
\end{equation}

The unsupervised objective is essentially identical to the CUDA objective and is defined as a joint log density over the labels and model parameters as follows:
\begin{equation}
\text{log } q ({\bf x}^t, y^t | \theta) = \text{log } \left[  \frac{p (y^t | {\bf x}^t, \theta) }{\sum_{i=1}^N  p (y^t | {\bf x}_i^t, \theta)} p(y^t) \right]
\label{equn:q-density}
\end{equation}

This joint distribution seeks to reduce the model's predictive bias when faced with a distribution shift in the target domain (see Fig.~\ref{figure:introduction}B). The denominator, which reflects the total probability of predicting a particular class, normalizes for biases originating from the source data. In other words, model predictions are normalized as $p (y^t | {\bf x}^t, \theta) /\sum_{i=1}^N  p (y^t | {\bf x}_i^t, \theta)$, such that no class has a biased probability of being predicted. Finally, the model predictions align with the target label distribution, using the supplied marginal $p(y^t)$.

Note, that this objective does not depend on whether the samples are labeled or not. Optimizing the joint distribution:
1) normalizes inherent bias towards one label over the other by calibrating the prediction for a label with its total weight in the dataset, 
2) and explicitly trains $q$ to match the marginals -- i.e., $\sum_{i=1}^N  q ({\bf x}_i^t, y^t | \theta) = p(y^t) $, thereby matching high-level statistics like per-class proportions in the training distribution. Importantly, with an informative prior $p(y^t)$ that specifies the putative distribution of the target labels, pseudo-label prediction can be biased toward labels that are more likely (see Section~\ref{subsection: model optimization}).

Combining the supervised and the unsupervised objective, the semi-supervised CRAFT objective is defined as:
\begin{equation}
\mathcal{L}(\mathcal{D}^t, \theta) = \sum_{i=1}^{N_l} \text{log } p (y_i^t | {\bf x}_i^t, \theta)  + \alpha \sum_{i=1}^N \text{log } q ({\bf x}_i^t, y_i^t | \theta)
\label{equn:craft-objective}
\end{equation}
In the above equation, $\alpha$ is the weight associated with the unsupervised objective and is tuned as a hyperparameter on a held-out cross-validation set during training. For deep-transfer, model parameters are initialized to $\theta^s$ derived from the source model, and then gradient descent is performed to obtain $\theta^t$.

\noindent{\bf Theoretical motivation}. Artificial Neural Networks (ANNs) with hypotheses space ${\bf \Theta}$ are trained to learn a parametric approximation $p({\bf x}, y| \theta)$, $\theta \in {\bf \Theta}$, such that the log-likelihood of the training data given the parameters is maximized. However, training ANNs often does not account for prior distributions over model parameters. In Appendix~\ref{appendix: MAP estimation view}, we demonstrate that the CRAFT semi-supervised objective can be theoretically derived as MAP estimation of the parameters ($ \text{log } p (\theta | \mathcal{D})$), which we write as a sum of the supervised objective, $\text{log } p (\mathcal{D} | \theta$), and a prior on the model parameters, $p (\theta)$. Specifically, we demonstrate, using the principle of entropy maximization \cite{Grandvalet2004}, that the unsupervised objective is related to the prior on the model parameters. 
\subsection{Model optimization}
\label{subsection: model optimization}
We optimize the supervised loss of the CRAFT objective with conventional gradient descent, jointly optimizing both components in the loss (equation~\ref{equn:craft-objective}). The unsupervised component of the loss is dependent on the labels $y^t$, which are unavailable for the vast majority of target data. To tackle this, we employ a 2-step optimization process.




\noindent{\bf Pseudo-label selection}. As the first step, we fix the model parameters $\theta$ and select the label that maximizes the joint distribution for each datapoint:
\begin{equation}
\tilde{y}_i^t = \underset{y^t \in \mathcal{Y}}{\text{argmax  }} q({\bf x}_i^t, y^t | \theta)
\label{equn:pseudo-label}
\end{equation}
where $\tilde{y}_i$ the pseudo-label for feature vector ${\bf x}_i$. 
For classification tasks, the optimization in equation \ref{equn:pseudo-label} can be achieved by selecting the best label from a finite set of class labels that maximizes $q$ \cite{Balgi2019}. But, for regression problems such as ours, class labels map to real numbers spanning the range of the model outputs $f(.; \theta): \mathbb{R}^d \mapsto [a, b]$. To address this, we rewrite the above equation under the assumption that the continuous-valued labels are distributed normally, with mean $f({\bf x}^t; \theta)$, and variance $c$. Hence, the above equation can be written as: 
\begin{equation}
\tilde{y}_i^t = \underset{y^t \in \mathcal{Y}}{\text{argmax  }} \frac{\mathcal{N} (y^t; f({\bf x}_i^t; \theta), c) p(y^t)}{\sum_{l=1}^N \mathcal{N} (y^t; f({\bf x}_l^t; \theta), c)}
\label{equn:regression-pseudo-label}
\end{equation} 

\begin{figure*}[!h]

\begin{center}
\includegraphics[scale=0.52]{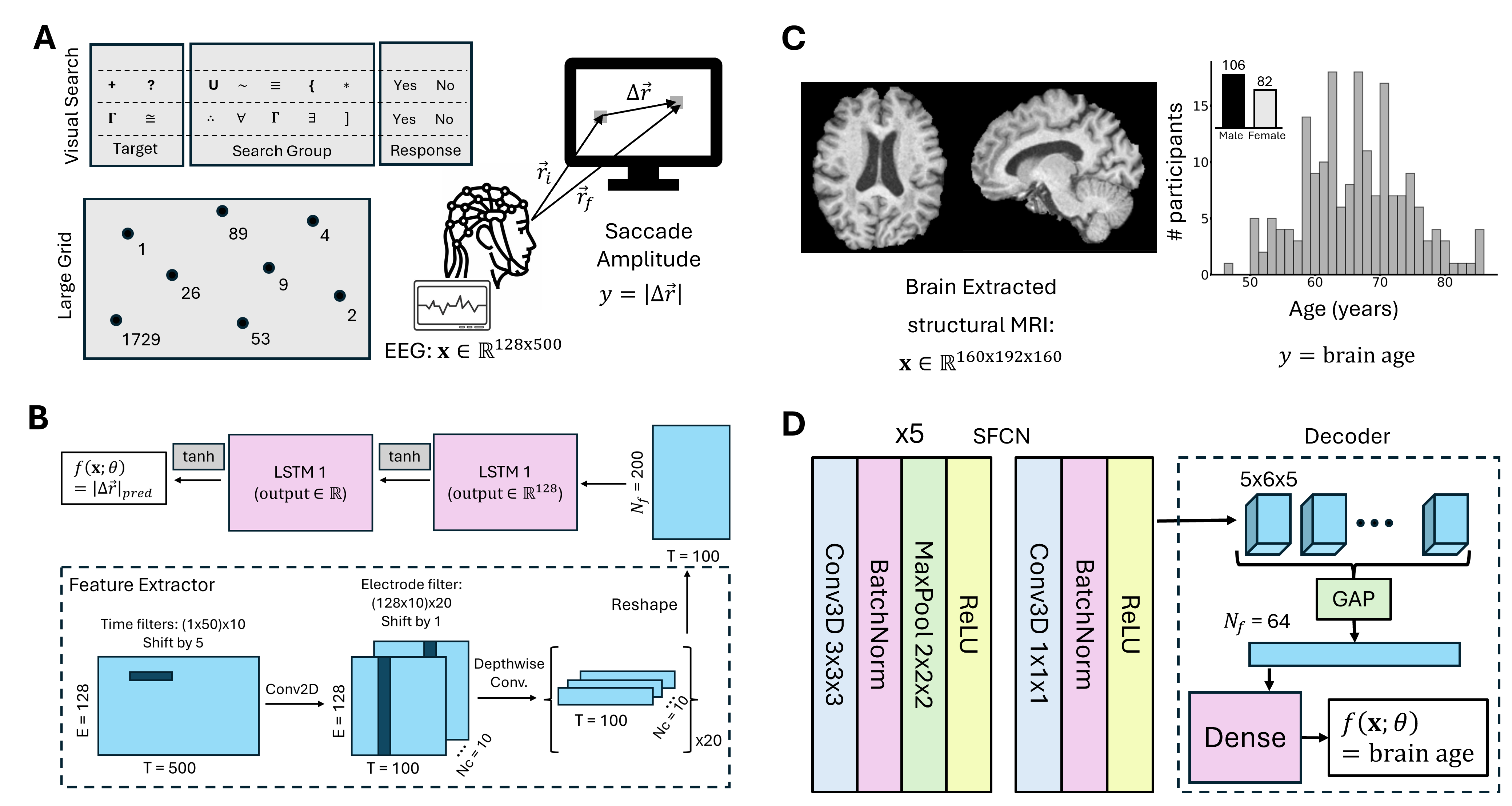}
\end{center}
\caption{({\bf A}) Schematic of gaze prediction using EEG signals. (Left) The task schematic of the Visual Search and Large Grid datasets from the EEGEyeNet benchmark. (Right) EEG and eye tracking data were recorded~\cite{Kastrati2021}. ({\bf B}) Architecture of the EEGNet-LSTM model, used to decode from EEG data. The parsimonious architecture can be used to decode saccades from EEG data. ({\bf C}) Schematic of the ``brain age'' prediction task using structural MRI scans. (Left) A sample preprocessed structural MRI scan by SFCN. (right) The age and gender (inset) distribution of the TLSA dataset. ({\bf D}) Model architecture of the SFCN-based decoder. The SFCN backbone was used to extract features from the MRI scan, which were used by a decoder block to predict brain age.}
\label{figure:Experiments}
\vspace{-0.4cm}
\end{figure*}

Equation \ref{equn:regression-pseudo-label} can be optimized with gradient ascent with respect to $y^t$ for the objective $\text{log } q({\bf x_i}^t, y^t | \theta)$. But, the pseudo-labels must be estimated before gradient descent can be performed on the model parameters (see next step, equation~\ref{equn:maximization}). This requires nested gradient descent, which increases computational complexity. Moreover, to employ an informative prior over target labels, we fit a class of mixture models (described in Appendix \ref{appendix: EMM + GMM}) to estimate $p(y)$ from the data; this makes backpropagation more cumbersome. To overcome these challenges, we divide the entire range of labels into small, discrete bins whose midpoints are the candidate pseudo-labels. Hence, this step in the optimization can be achieved faster pseudo-label selection from a discrete set. Note that this approach does not transform the model outputs to be discrete; it merely represents a convenient optimization strategy to update the parameters. In the experimental section, we provide a recipe for choosing bin sizes and study the impact of varying bin sizes. 


\noindent{\bf Maximization.}
In the second step, we optimize the model parameters by using the labeled data $\{{\bf x}_i^t, y_i^t\}_{i=1}^{N_l}$ and fixing the pseudo-labels $\{{\bf x}_i^t, \tilde{y}_i^t\}_{i=1}^N$, as:

$$ \theta^* = \underset{\theta \in {\bf \Theta}}{\text{argmax  }}  \sum_{i=1}^{N_l} \text{log } p (y_i^t | {\bf x}_i^t, \theta)  + \alpha \sum_{i=1}^N \text{log } q ({\bf x}_i^t, \tilde{y}_i^t | \theta) $$

After incorporating distributional assumptions in the above equation and subsuming constants with respect to model parameters (e.g., priors on the labels $p(y^t)$):
\begin{multline}
\theta^* = \underset{\theta \in {\bf \Theta}}{\text{argmax  }}  - \sum_{i=1}^{N_l}  (y_i^t - f({\bf x}_i^t; \theta))^2 - \alpha(\sum_{i=1}^N  \left(\tilde{y}_i^t - f({\bf x}_i^t; \theta)\right)^2   \\ 
                                              - \sum_{i=1}^N  \text {log } \sum_{l=1}^N \text{exp } (- (\tilde{y}_i^t - f({\bf x}_l^t; \theta))^2))    
\label{equn:maximization}
\end{multline}

In practice, each update is performed with a batch of labeled and unlabeled data. First, pseudo-labels are computed for the unlabeled data. Then, keeping the pseudo-labels fixed, the second and third terms of this objective are optimized, while the labeled data are used to optimize the first term, concurrently.  Note that the optimization is performed with the model parameters initialized to $\theta^s$, derived from the source model.


Intuitively, the supervised loss in Equation \ref{equn:maximization} encourages model predictions to align with their true labels; because model parameters were initialized with the source model, this acts as an implicit regularizer discouraging the adapted model from deviating too far from the source model. The second term forces dissimilar predictions for data points with different pseudo-labels. In other words, providing samples with dissimilar label values during training produces disparate representations for data points with distant labels, which in turn helps to learn a better regression line. We use the log-sum-exponent trick to avoid numerical instability during optimizing of the second term.

\section{Datasets and Competing Methods}
We employ CRAFT to address challenges relating to data scarcity, missing labels, and continuous label prediction in neuroscience.

\subsection{Datasets and models}
\subsubsection{Gaze prediction from EEG signals} 
\label{subsubsection: Datasets EEG}
Saccades are rapid ballistic eye movements, putatively linked to higher-order cognitive functions like attention and working memory~\cite{Deubel1996, Brincat2021}.  Decoding eye movements from neural data could help neuroscientists develop better models of these complex processes. With this objective, the EEGEyenet benchmark dataset~\cite{Kastrati2021} was collected from $356$ participants. It contains $\sim 47.5$ hours of 128-channel electroencephalogram (EEG) recordings with eye-tracking data collected concurrently.

Here, we predict saccade amplitudes from these EEG data. We consider two datasets -- the ``Large Grid'' and the ``Visual Search'' datasets -- from the EEGEyeNet benchmark~\cite{Kastrati2021}  (Fig.~\ref{figure:Experiments}A). In the Large Grid paradigm, participants made saccades to a target location cued by a dot on the screen. In the Visual Search paradigm, participants searched for a specific symbol in the display and, thus, made saccades during the search.

\noindent {\bf Preprocessing}.
Each dataset was preprocessed using the pipeline described in the benchmark (details in Appendix~\ref{appendix: Datasets EEG}). We decode saccade magnitudes ($y \in \mathbb{R}^+$) using the 128-channel EEG timeseries, sampled at 500Hz (${\bf x} \in \mathbb{R}^{128\text{x}500}$), for corresponding trials. Table~\ref{table:eye-datasets} (see Appendix~\ref{appendix: Datasets EEG}) lists the details of the train, cross-validation and test distributions for these datasets. The Visual Search paradigm has more training data and was used to train the source model; this was then transferred to predict saccade amplitudes in the Large Grid dataset (target). 

\noindent {\bf Models}.
The EEGEyeNet benchmark showed that Pyramidal CNNs \cite{Ullah2016} were most effective at predicting saccade amplitude. As an alternative to 2D convolutional networks, we also evaluate the performance of a novel end-to-end Long-Short-Term-Memory (LSTM) model. This model is trained on features extracted by separable convolutions along time and electrode dimensions, following EEGNet~\cite{Lawhern2018} (Fig.~\ref{figure:Experiments}B, more details in the Appendix~\ref{appendix: Datasets EEG}). 

\subsubsection{``Brain age'' prediction from MRI scans}
\label{subsubsection: Datasets Brain Age}
``Brain age'' is the biological age of the brain and is an important biomarker for cognitive health. Studies have shown that accelerated brain aging is associated with serious cognitive impairments like Alzheimer's disease~\cite{Cole2018}. Hence, estimating brain age is a critical problem in neuro-medicine. 

As a first step, attempts have been made to decode the chronological age of healthy participants using T1-weighted structural MRI scans~\cite{Peng2021}. A 3D-CNN model (see next, Models) was trained using the large UK-Biobank dataset (N=14,503, 44-80 years, mean age=52.7 years)~\cite{Miller2016}. We seek to transfer this pretrained model to decode age in a small Indian cohort -- the TATA Longitudinal Study of Aging (TLSA) (N=188, 46-88 years, mean age = 66 years)~\cite{Sundarakumar2022} (Fig.~\ref{figure:Experiments}C), without access to the UKBiobank data. 


\noindent {\bf Preprocessing}.
We followed a preprocessing pipeline identical to~\cite{Peng2021}. The T1w-MRI scan was brain-extracted, bias-corrected~\cite{Tournier2019}, and registered to standard MNI-152 space~\cite{Mandal2012}, and cropped to get preprocessed scans ${\bf x} \in \mathbb{R}^\text{160x192x160}$, which was used to predict brain age. 

\noindent {\bf Models}.
Features from the pretrained Simple Fully Convolutional Network (SFCN, Fig.~\ref{figure:Experiments}D) were used to make the predictions with a regression model; more details are provided in Appendix~\ref{appendix: The SFCN Architecture})~\cite{Peng2021}. The entire model was then adapted to the target dataset (TLSA). 

We also apply CRAFT to people counting and tumor prediction tasks (see Appendix~\ref{appendix: People Counting Methods},~\ref{appendix: tumor-size prediction data} for dataset details).


\subsection{Competing DA Approaches}
\label{subsection: Competing Methods}
\noindent {\bf Baseline model}. We define a naive baseline model that always predicts the mean label of the training set, ignoring the input features of the test samples.

\noindent {\bf Transfer Learning (TL)}.  
We naively adapt the source model by minimizing the supervised loss on the labeled samples in the target dataset. 

\noindent{\bf TASFAR}. 
This is an SF-UDA model (See Section~\ref{section: related works}) for regression problems. It depends on two statistics from the source dataset -- the threshold for confident samples and the mapping between uncertainty to prediction errors. Because the source is unavailable, we compute these using the TL model on the target. For a fair comparison, we add the supervised loss on the labeled data to the unsupervised loss, weighted by the same $\alpha$ as in our model.

\noindent{\bf DataFree}. This is another SF-UDA model (See Section~\ref{section: related works}). The original model depends on the distribution of intermediate latents on the source distribution. Again, because the source data is unavailable, we computed latent distribution using the TL model and the labeled target data. As before, the objective was modified to accommodate a supervised loss (same $\alpha$ as in our model). Bin sizes for both TASFAR and DataFree were selected to be the same as in CRAFT.  

\noindent{\bf Progressive Mixup}. This is a semi-supervised, source-free domain adaptation tool (See Section~\ref{section: related works}). We adapt the model to the regression task by replacing cross-entropy losses with mean-squared error loss. Predictive entropy was replaced by variance to estimate uncertainty.  

\noindent{\bf BBCN}. Bilateral Cycle Consistency is a prototype-based clustering SF-SSDA paradigm (See Section~\ref{section: related works}). It depends on class-level prototypes for each model. To extend it to regression, we discretize the output space into bins of the same width as the other methods.



\noindent{\bf Metrics. } We quantify the prediction performance of continuous valued outputs using two metrics -- Root Mean Squared Error (RMSE) and Percentage-Bend Correlation~\cite{Wilcox1994} coefficient (R). A {\it lower} RMSE reflects a closer correspondence between the predicted and actual labels, a {\it higher} R-value quantifies better predictions along the best-fit line between the actual and predicted labels. 

In subsequent sections (\ref{subsection: Saccade Amplitude prediction from EEG} and~\ref{subsection: Brain Age Prediction from structural MRI}), we predict saccade amplitude from EEG and brain age from structural MRI scans, respectively.  



\section{Experiments: Results}
\label{section: Experiments}

\subsection{Saccade Amplitude Prediction}
\label{subsection: Saccade Amplitude prediction from EEG}

Firstly, we trained an EEGNet-LSTM model on the Visual Search source dataset (RMSE=67.03, R=0.91). This novel architecture outperformed the benchmark model (Pyramidal CNNs) by $\sim 16\%$ (benchmark: RMSE=79.59 pix, R=0.87). Next, this was transferred to the Large Grid dataset. The upper 3 rows of Table~\ref{table: Saccade sf-ssda} shows the performance when all the data was used for funetuning. In this case also EEGNet-LSTM outperformed the benchmark (Pyramidal CNN~\cite{Kastrati2021}) by a large margin ($\sim19\%$). Hence, subsequent SSDA experiments were performed using EEGNet-LSTMs. 

All models were trained with mini-batches of 128 samples, and only the best-performing checkpoint on the held-out validation set was used for evaluating the test set. Adam optimizer with an initial learning rate of $10^{-4}$ was used. $\alpha$ in equation~\ref{equn:craft-objective} was selected with a grid saerch over $\{0.01, 0.1, 1.0\}$; empirically, $\alpha=0.1$ was always favored.

\begin{table}[!h]
\caption{
Source Free Semi-supervised Deep Transfer of the model trained on the Visual Search task to the Large-Grid Task. The upper part of the table (rows 1-3) shows the performance ceiling when all labeled data in the target is used for training. The lower part (rows 4-9) shows SF-SSDA results when only 1\% of the target samples were labeled (3 seeds). 
}
\label{table: Saccade sf-ssda}
\centering
\resizebox{1.01\linewidth}{!}{%
\begin{tabular}{lcc}
\toprule 
{\textbf{Method}} & \textbf{R} $\uparrow$ & \textbf{RMSE } (in pix) $\downarrow$\\ \midrule
Naive Baseline                      &            -           &   149.12 $\pm$ 0.02   \\
Benchmark  (TL, 100\%)  &    0.91 $\pm$ 0.01     &   63.84 $\pm$ 0.75    \\
EEGNet-LSTM (TL, 100\%)          &    \bf{0.93 $\pm$ 0.01}     &   \bf{51.47 $\pm$ 0.63}   \\
\hline
{\it SF-SSDA} (1\% labels) \\
EEGNet-LSTM + TL           &    0.77 $\pm$ 0.01        &   92.26 $\pm$ 1.66    \\
EEGNet-LSTM + Mixup        &    0.48 $\pm$ 0.02        &   135.70 $\pm$ 1.25     \\
EEGNet-LSTM + BBCN         &    0.76 $\pm$ 0.01        &   99.8 $\pm$ 3.35     \\
EEGNet-LSTM + TASFAR       &    0.76 $\pm$ 0.01        &   \it{86.41 $\pm$ 1.05}    \\
EEGNet-LSTM + DataFree     &    \it{0.80 $\pm$ 0.01}        &   87.64 $\pm$ 3.08    \\
EEGNet-LSTM + $\textcolor{red}{\text{CRAFT}}$    &  \bf{0.81 $\pm$ 0.02}  &  \bf{84.17 $\pm$ 3.95}   \\
\bottomrule
\end{tabular} %
}
\end{table}

To demonstrate the efficacy of CRAFT for SF-SSDA, target labels from the training data of the Large Grid dataset were randomly dropped in a stratified manner. Figure~\ref{figure:semisup-all, main paper}A shows how the performance of these models is affected by reducing the fraction of labeled data ($n_{ul}/N$ is the fraction of unlabeled samples). A lower proportion of labeled data worsens the performance of all the models (RMSE increases), but CRAFT produces the lowest RMSE for all unlabeled data proportions. Figure~\ref{figure:semisup-all, main paper}B shows the prediction of saccade magnitudes ($|\Delta \overline{r}|$) for the best CRAFT-based EEGNet-LSTM model. Table~\ref{table: Saccade sf-ssda} (lower 6 rows) summarizes the performance of CRAFT when 99\% of the data were unlabeled;  CRAFT outperforms supervised training (TL) by $9 \%$ and other SOTA SF-SSDA models by $> 4\%$ in terms of RMSE. The closest competitive methods were TASFAR and DataFree. The correlation coefficient (R) also showed similar trends (Table~\ref{table: Saccade sf-ssda}; Fig.~\ref{figure: All Rs}A). 


\subsection{Brain Age Prediction}
\label{subsection: Brain Age Prediction from structural MRI}
\label{subsubsection: Performance Brain Age}
Next, we transferred the SFCN model trained on the UK Biobank MRI scans to the TLSA dataset. The TLSA dataset has only 188 MRI scans, so we avoid hyperparameter searches. All models were trained for 25 epochs (batch size = 4) using the Adam optimizer (initial learning rate = $10^{-4}$), and only the final model was used for inference. Here, we used $\alpha = 0.1$; we explore the effect of varying $\alpha$ for ours and competing models in Appendix~\ref{Appendix: More Hyperparameter details}. Each metric was calculated for the prediction on the entire dataset, leaving out one fold at a time (4-folds). 
The top of Table \ref{table:Brain Age Semisup} (row 2) quantifies the performance ceiling when all the training data are used (3 seeds).

\begin{table}[!h]
\caption{
Semi-supervised transfer of the SFCN model (source: UKB) to the TLSA dataset. Same as in Table 1, except that the lower part of the table (3-8) shows SF-SSDA results when only 20\% of the target samples were labeled. (3 seeds)
}
\label{table:Brain Age Semisup}
\centering
\resizebox{1.01\linewidth}{!}{%
\begin{tabular}{lcc}
\toprule 
{\textbf{Method}} & \textbf{R} $\uparrow$ & \textbf{RMSE } (in years)  $\downarrow$\\ \midrule
Naive Baseline                  &           -                  &   7.91 $\pm$ 0.05      \\
SFCN + TL (100\%)               &        0.66 $\pm$ 0.01       &    6.14 $\pm$ 0.03     \\
\hline
{\it SF-SSDA} \text{(20\% labels)} \\
SFCN + TL            &    0.41 $\pm$ 0.07      &   7.41 $\pm$ 0.21      \\
SFCN + Mixup         &    0.34 $\pm$ 0.04      &   7.71 $\pm$ 0.14      \\
SFCN + BBCN          &    0.28 $\pm$ 0.04      &   8.00 $\pm$ 0.15      \\
SFCN + TASFAR        &    0.42 $\pm$ 0.07      &   7.47 $\pm$ 0.15      \\
SFCN + DataFree      &    \it{0.50 $\pm$ 0.03}      &   \it{7.36 $\pm$ 0.14}      \\
SFCN + $\textcolor{red}{\text{CRAFT}}$   &  \bf{0.51 $\pm$ 0.03}  &  \bf{7.14 $\pm$ 0.11}  \\
\bottomrule
\end{tabular} %
}
\end{table}

For SF-SSDA, we performed experiments closely similar to the saccade task (Section~\ref{subsection: Saccade Amplitude prediction from EEG}). Figure~\ref{figure:semisup-all, main paper}C shows how these models' performance is affected when the proportion of unlabeled samples increases. Reducing the fraction of labeled data worsens model performance, but CRAFT continues to remain the most competitive of all models. Figure~\ref{figure:semisup-all, main paper}D shows the prediction scatter of the best CRAFT-based model. Table~\ref{table:Brain Age Semisup} summarizes the performance when 80\% of the data was unlabeled. CRAFT outperforms supervised training (TL) by $\sim 4 \%$ and other SOTA SF-SSDA models by $> 3\%$ in terms of RMSE. Similar results were obtained with the R values also (Fig.~\ref{figure: All Rs}B). Again, TASFAR and DataFree emerge as the most competitive methods.  

\begin{figure}[!h]
\begin{center}
\includegraphics[scale=0.33]{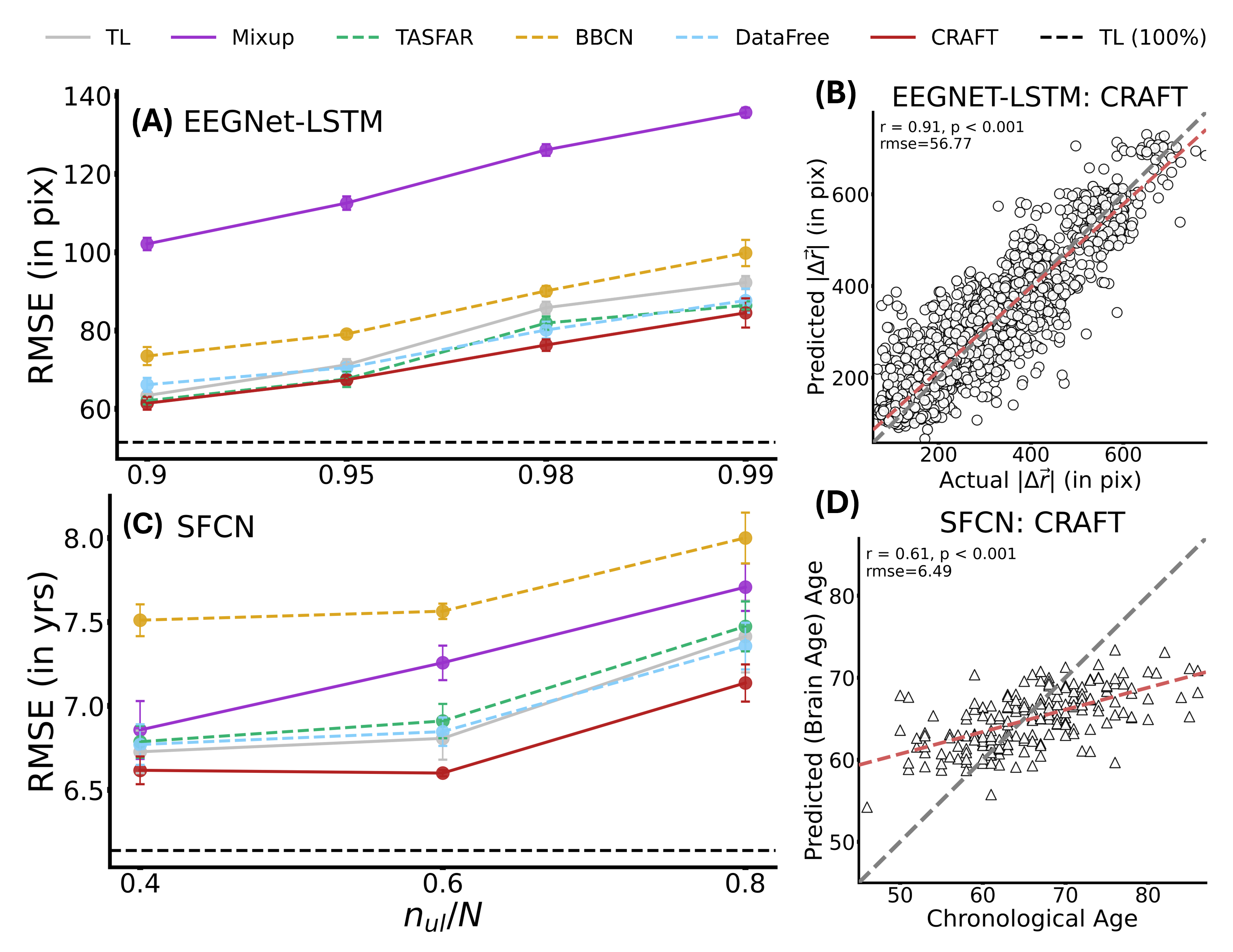}
\end{center}
\caption{({\bf A, C}) RMSE ($\downarrow$) for source-free semi-supervised domain adaptation of saccade ({\bf A}) and brain age ({\bf C}) prediction tasks, trained with varying proportions of unlabeled target data $n_{ul}/N$. Dashed black line: Performance ceiling. Circles: individual trials. ({\bf B, D}) Prediction for the best CRAFT model for saccade amplitude ({\bf B}) and brain age ({\bf D}), with 90\% and 40\% unlabeled data, respectively. Triangles: individual participants. Results with a much higher proportion of unlabeled data are shown for the saccade dataset because of the far greater size of this ($N\sim$12k) compared to the brain age ($N\sim$180) dataset.
}
\label{figure:semisup-all, main paper}
\end{figure}

Finally, we evaluated the effectiveness of CRAFT at two other real-world regression benchmarks: People Counting $~\cite{JHU, gao2020nwpu}$ and Tumor Size prediction $\cite{cam16, cam17}$.
CRAFT outperformed other SF-SSDA methods by $> 5\%$ and $> 2\%$ respectively for each dataset; the details are presented in Appendices~\ref{appendix: people counting} and ~\ref{appendix: tumor-size}.

\noindent{\bf Mitigating Sampling Biases. } Next, we show how the unsupervised CRAFT objective could mitigate sampling bias effects in the target training set. In all of the previous simulations, the label distributions of the target train and test data were matched. Here, we deliberately distorted the label distribution of the target training data: 80\% of the data above the mean age in the training set were removed, creating a heavy bias toward lower ages. Following this, labels were retained in only 40\% of the training data ($n_{ul}/N=0.6)$. No such biases were introduced into the test data. The CRAFT model was then trained by imposing the unbiased (true) marginal label distribution on the unlabeled training set (equation~\ref{equn:regression-pseudo-label}). In this scenario, again, we find that CRAFT mitigates the effect of sampling biases in the training data (Table \ref{table:sampling-error}). It outperforms transfer learning by $\sim 5\%$, and other sota SF-SSDA models by $> 2.5\%$ (RMSE).

\begin{table}[!h]
\caption{
Analyses with datasets with sampling error. 80\% of the data above the mean age is removed. Furthermore, only 40\% of the labels in this mislabeled data are used for semi-supervised transfer of the SFCN model (source: UKB). (3 seeds) 
}
\label{table:sampling-error}
\centering
\resizebox{1.01\linewidth}{!}{%
\begin{tabular}{lcc}
\toprule 
{\textbf{Method}} & \textbf{R} $\uparrow$ & \textbf{RMSE } (in years)  $\downarrow$\\ \midrule

Naive Baseline       &           -             &   8.38 $\pm$ 0.22      \\
SFCN + TL            &    0.41 $\pm$ 0.05      &   8.40 $\pm$ 0.27      \\
SFCN + Mixup         &    0.32 $\pm$ 0.02      &   \it{8.19 $\pm$ 0.15}      \\
SFCN + BBCN          &    0.23 $\pm$ 0.02      &   8.53 $\pm$ 0.16      \\
SFCN + TASFAR        &    0.39 $\pm$ 0.04      &   8.57 $\pm$ 0.22      \\
SFCN + DataFree      &    \it{0.41 $\pm$ 0.03}      &   8.58 $\pm$ 0.23      \\
SFCN + $\textcolor{red}{\text{CRAFT}}$   &  \bf{0.45 $\pm$ 0.03}  &  \bf{7.98 $\pm$ 0.25}  \\
\bottomrule
\end{tabular} %
}
\end{table}



\subsection{Analysis of computational complexity}
\label{subsection: Computational Complexity}
Table~\ref{table:complexity} shows the computational complexity of CRAFT and the competing models. We compute both the algorithms' theoretical complexity and running time on the saccade and brain age prediction tasks. The theoretical training complexity is defined as the number of gradient updates per epoch. For inference, the complexity is the same for all models as the same architecture is trained using different adaptation paradigms. Training time and complexity are lowest for TASFAR, amongst all other SF-DA tools. The complexity of CRAFT was comparable to DataFree, the next best method.

\begin{table}[h!]
\caption{
{\bf Computational complexity}. Time per training epoch and for all test data (in {\bf min}). Parentheses: P- no. of model parameters; N- samples; D- latent dimensionality; $\beta$- batch size, B- no. of bins, m- no. of inferences for MC Dropout uncertainty estimation.}
\label{table:complexity}
\centering
\resizebox{1.0\linewidth}{!}{%
\begin{tabular}{lcccc}
\toprule 
 & \multicolumn{2}{c}{\it \textcolor{gray}{\bf EEGNet-LSTM} (P $\sim$ 190K)}  & \multicolumn{2}{c}{\it \textcolor{gray}{\bf SFCN} (P $\sim$ 2.95M)} \\ 
{(B) \textbf{Method}} & \textbf{Train /Comp.} $\downarrow$ & \textbf{Test /Comp.} $\downarrow$  &  \textbf{Train/Comp.} $\downarrow$  & \textbf{Test/Comp.} $\downarrow$  \\
& (N $\sim$ 12k) &  (N $\sim$ 2.5k) &  (N $\sim$ 150) & (N $\sim$ 50) \\
\midrule
TL                                      &   0.15 / \it{O}(N)  &   0.02 / \it{O}(N)   &   0.12  / \it{O}(N)  &    0.04 / \it{O}(N)  \\
Mixup                                   &   0.74 / \it{O}(N$m$)  &   0.02 / \it{O}(N)   &   0.74  / \it{O}(N$m$)  &    0.04 / \it{O}(N)  \\
BBCN                                    &   2.71 / \it{O}(N$BD$)  &   0.02 / \it{O}(N)   &   1.29  / \it{O}(N$BD$)  &    0.04 / \it{O}(N)  \\
TASFAR              &   0.30 / \it{O}(N)  &   0.02 / \it{O}(N)   &   0.19 / \it{O}(N)  &    0.04 / \it{O}(N)   \\
DataFree           &   0.55 / \it{O}(N$D^2$)  &   0.02 / \it{O}(N)   &   0.32 / \it{O}(N$D^2$)   &    0.04 / \it{O}(N)   \\
\textcolor{red}{CRAFT}      &   0.45 / \it{O}(N$(\beta + B)$)  &   0.02 / \it{O}(N)   &   0.36 / \it{O}(N$(\beta + B)$)   &    0.04 / \it{O}(N)   \\
\bottomrule
\end{tabular} %
}
\end{table}

\section{Concluding Remarks}
\label{section: Discussion}

Data scarcity poses a serious challenge for applying deep-learning models in real-world settings. Training over-parameterized models on a few hundred data points leads to overfitting and poor generalization. Conventional domain adaptation algorithms, to transfer pretrained networks, typically need access to source domain data, but these may not always be available. In this study, we propose a source-free semi-supervised domain adaptation framework that recalibrates the decision boundary without aligning the intermediate representation of the domains. 
In fields like medicine, resource constraints associated with labeling data are significant; therefore, leveraging the large amounts of unlabeled data becomes critical.
We propose CRAFT as an SF-SSDA approach for such source-free, label-sparse settings that are common in real-world applications.


\section{Acknowledgements}
This work was supported by the following funding sources: a Prime Minister’s Research Fellowship (to M.B.), a DST SwarnaJayanti fellowship, a Pratiksha Trust Intramural grant, a Tata Trusts grant and a Google Research grant (to D.S.). We also thank the Tata Longitudinal Study on Aging at the Indian Institute of Science for access to MRI scans from their cohort.
\clearpage
{
    \small
    \bibliographystyle{ieeenat_fullname}
    \bibliography{main}
}

\clearpage
\clearpage
\appendix
\setcounter{page}{1}
\maketitlesupplementary
\section{Derivations}

\subsection{CRAFT: A MAP Estimation View}
\label{appendix: MAP estimation view}

In this section, we show that the proposed method can be viewed as regularized training of neural networks. In particular, we show that optimization of the CRAFT objective gives us the maximum a posteriori (MAP) estimate of model parameters. We know the MAP estimate for the parameters $\theta$ (given dataset $\mathcal{D}$) is obtained by maximizing the posterior $ \text{log } p (\theta | \mathcal{D}) \propto \text{log } p (\mathcal{D} | \theta)  +  \text{log } p (\theta) $, where, $\text{log } p ( \mathcal{D} | {\bf \theta})$ is the log-likelihood, and $p (\theta)$ is the prior distribution (regularizer) over the model parameters. The subsections below describe the choices for these components.

\subsubsection{Likelihood}
\label{subsection:Likelihood}

We define the log-likelihood for a labeled data as: $\text{log } p (\mathcal{D} | \theta) = \sum_{i=1}^N \text{log } p (y_i | {\bf x}_i, \theta)$. As this paper deals with unlabeled data, as a design choice, we assume the conditional distribution $p (y | {\bf x}, \theta)$ to be constant for unlabeled training data, optimizing the log-likelihood over labeled data only. 

Although we only address continuous label prediction (regression -- $p (y | {\bf x}, \theta) = \mathcal{N} (y; f({\bf x}; \theta), c)$) tasks in this study, the framework can be converted to k-way discrete label prediction (classification) tasks by appropriate assumptions on the conditional distribution -- e.g., $p (y | {\bf x}, \theta) = \text{Mult}_k (y; f_1({\bf x}; \theta),..., f_k({\bf x}; \theta))$.

\subsubsection{The Prior Distribution}
\label{subsection: The Prior Distribution}

In this subsection, we derive a principled prior distribution that can leverage unlabeled data and/or deal with data scarcity during transfer learning by biasing the selection of parameters that best suit the domain. We achieve this by extending the CUDA~\cite{Balgi2019} framework, originally used for domain adaptation.

We note a few desirable properties for supervised transfer learning algorithms. Firstly, these models are not explicitly trained to match the marginal label distribution, i.e., $p(y) \neq \int_{\bf x} p(y | {\bf x}, \theta) p({\bf x}) d{\bf x}$. Hence, the model's performance on unseen data can be compromised if the training and the test label distributions differ, such as due to sampling biases. Hence, $p(y) =  p(y | \theta)$ is a property we desire. Secondly, ensuring this property can be particularly useful if we know that the training dataset is small, where sampling biases are likely. Hence, we define a joint distribution using the trained model $p (y | {\bf x}, \theta) $ to tackle these challenges: 

\begin{equation}
\text q ({\bf x}, y | \theta) = \left[  \frac{p (y | {\bf x}, \theta) }{\sum_{i=1}^N  p (y | {\bf x}_i, \theta)} p(y) \right]
\end{equation}

Now, we incorporate $q ({\bf x}, y | \theta)$ as a prior over the model parameters. We use the maximum entropy principle to achieve this. In particular, we wish to maximize the entropy of the model's parameter distribution such that the negative log-probability of the joint distribution $q$, on average is as small as a constant $G$, i.e., $\mathbb{E}_{\theta \in {\bf \Theta}} [- \text{log } q ({\bf x}, y | \theta)] = G$. Solving the Euler-Lagrange equations we get (see Appendix~\ref{appendix: The Prior Distribution}):
\begin{equation}
\text{log } p(\theta) \propto {\alpha \text{log } q ({\bf x}, y | \theta)}
\label{equn:prior-density}
\end{equation}
where $\alpha$ is the Lagrange multiplier corresponding to $G$.

\subsection{A Maximum Entropy Prior}
\label{appendix: The Prior Distribution}


The optimization objective described above can be expressed as:

\begin{equation}
\theta^* =  \underset{\theta \in \Theta}{\text{argmax}} - \int_{\theta} {p(\theta) \text{ log } p(\theta)} \ud \theta 
\label{equation: constrained CDIST}
\end{equation}
$$\text{s.t.} \quad \mathbb{E}_{\theta \in {\bf \Theta}} [- \text{log } q ({\bf x}, y | \theta)] = G, \int_\theta p(\theta) \ud \theta = 1 
$$

We ignore the constraint on the density integral,  and normalize it at the end. Hence, the Lagrangian can be written as:

$$
\mathcal{L} (\theta) = \int_{\theta} \underbrace{ {p(\theta) \text{  } (-\text{log } p(\theta)}  + \alpha (\text{log } q ({\bf x}, y | \theta) + G)) \text{  } }_{g(\theta, \text{ } p(\theta))} \ud \theta
$$

Using the Euler-Lagrangian equation, and because $g$ does not depend on the first derivate of $p$:

$$
\frac{\partial g}{\partial p} - \frac{\ud}{\ud \theta} \frac{\partial g}{\partial p'} = -1 - \text{log } p(\theta)  + \alpha (\text{log } q ({\bf x}, y | \theta) + G) = 0
$$

$$
\implies \text{log } p(\theta) = \alpha \text{log } q ({\bf x}, y | \theta) +  (\alpha G - 1)  
$$

Dropping the constant terms that are independent of $\theta$:
$$
p(\theta) \propto {\alpha \text{log } q ({\bf x}, y | \theta)}.   \quad 
$$

\subsection{Mixtures of Gaussians and Exponentials}
\label{appendix: EMM + GMM}

For learning the marginal distribution of labels for CRAFT we fit a mixture of exponential and Gaussian distributions. Given samples $\{{\bf x}_i\}_{i=1}^N$ from an unknown distribution $p_x$, s.t., $x_i \in \mathbb{R}_+$. If negative components exist, we can add a constant offset to make the data non-negative. We model the underlying distribution by $k_1$ Gaussians and $k_2$ exponentials. The density of a mixture model $p_\theta$, can be written as;

\begin{equation}
p_\theta (x) =  \sum_{i=1}^{k_1 + k_2} p_\theta (z)  p_\theta (x | z)
\label{equation: EMM+GMM}
\end{equation}
where $z$ is the latent variable, s.t., $p_\theta (z) = \text{Mult}_{k_1 + k_2}(z; \beta_1, \beta_2, ..., \beta_{k_1 + k_2})$. The conditionals are defined as:

$$
p_\theta (x  | z=i) = \begin{cases} 
\mathcal{N} (x; \mu_i, \sigma_i^2) &  i \in \{1, 2, ..., k_1\} \\
\text{Exp}(x; \lambda_i) & i \in \{k_1 + 1, ..., k_1 + k_2\}
\end{cases}
$$

We estimate the parameters $\theta = \{ \{\beta_i\}_{i=1}^{k_1 + k_2} , \{\mu_i, \sigma_i^2 \}_{i=1}^{k_1}, \{\lambda_i \}_{i=k_1 + 1}^{k_2} \} $, using a custom designed expectation-maximization algorithm (code available in Appendix~\ref{appendix: code}).

\section{Datasets and Methods}

\subsection{Gaze prediction from brain signals}
\label{appendix: Datasets EEG}
{\bf Preprocessing.}
All eye-tracking datasets -- Visual Search and Large Grid -- were preprocessed similarly. Briefly, blinks were removed using Gaussian filtering, and the temporal locations of saccades were found by detecting very high velocity and acceleration. Then, the eye-tracking data was partitioned into 1s windows so that only one saccade event occurred in that window. The displacement vector of the eye was computed, and its magnitude (saccade amplitude) was designated the target variable $y \in \mathbb{R}^+$. Time-matched to each saccade, a 500Hz EEG signal was extracted and used as the feature vector ${\bf x} \in \mathbb{R}^{128\text{x}500}$ (128 electrodes or channels). Table \ref{table:eye-datasets} lists the details of the train, cross-validation, and test distributions for these datasets. We followed the leave-participant-out evaluation strategy with an identical train-validation-test split for the Large grid dataset for direct comparison with the benchmark. The Visual Search paradigm has more training data and was used to train the source model, which was transferred to the target datasets - Large Grid (from the benchmark). Before training the model, the target variable (saccade amplitude) was linearly scaled between $[-1, 1]$, and the EEG data were z-scored, by computing a common mean and standard deviation, across time and channels.

\begin{table}[!ht]
\caption{
Training, validation and test sets for each dataset. A 60-20-20\% leave-participant-out split was used, as in the original benchmark.
}
\label{table:eye-datasets}
\centering
\resizebox{\linewidth}{!}{
\begin{tabular}{lccc}
\toprule 

{\textbf{Dataset}} & \textbf{Train} & \textbf{Validation} & \textbf{Test} \\ \midrule

Visual Search (Source)       &  21,191  &   5,018   &  5,354  \\
Large Grid (Target)        &  12,275  &   2,836   &  2,719   \\
\bottomrule
\end{tabular} %
}
\end{table}

\noindent{\bf Models}. 
The EEG-EyeNet benchmark showed that Pyramidal CNNs~\cite{Ullah2016} performed best at predicting saccade amplitude. However, naive 2D convolution-based neural networks treat the time$\times$channel EEG data as an image, whereas such data typically lack the correlation structure associated with natural images. Moreover, CNN-based models are not ideal for learning temporal autocorrelation structure present in the data.

We address these shortcomings with a novel end-to-end Long-Short-Term-Memory (LSTM) model trained on features extracted by an EEGNet-like~\cite{Lawhern2018} architecture. Figure~\ref{figure:Experiments}B shows the feature extractor, which employs separable convolutions along time and electrode dimensions. The extracted temporal features are then fed to a series of two LSTMs, which predict the saccade amplitude. 

In the feature extractor, the EEG data is first convolved along the temporal dimension using 100 ms trainable filters independently for each electrode, as it is the average duration to execute a saccade \cite{Crouzet2010}. The filters are shifted by 10 ms to capture neural dynamics up to $\sim$100 Hz. Finally, depthwise-separable convolution is performed time-point-wise along the electrode dimension to obtain temporal features for the LSTM-based decoder.

\subsection{Brain age prediction from structural MRI scans}
\label{appendix: The SFCN Architecture}
{\bf Models}. 
The Simple Fully Convolutional Network (SFCN, Fig.~\ref{figure:Experiments}D)~\cite{Peng2021} feature Extractor contains five blocks of 3D Convolution (3x3x3 filters), Batch-Normalizaton, and 3D-MaxPooling (2x2x2 filters), and ReLU activation, followed by a single block of 3D Convolution with 1x1x1 filters, Batch-Normalizaton, and ReLU activation. The outputs of the feature extractor are then the Global Average Pooled (GAP) and fed as input to a classifier. In the original paper, the classifier predicted the probability of each MRI scan belonging to one of the 40 bins (range: 42-82 years). The expected value of the outputs was reported as the predicted age. In our version of SFCN, we avoid binning by replacing the classifier with a regressor.

\subsection{People counting from natural scene images}
\label{appendix: People Counting Methods}
We addressed the challenge of counting people from photographs of crowds -- the ``people counting'' challenge.
{\bf Preprocessing}. 
High-resolution scene images from two datasets --
NWPU ~\cite{gao2020nwpu} and JHU-crowd~\cite{JHU} -- were resized to 1152x768. Following this, each image was divided into six non-overlapping 384x384 patches. These patches were fed to a ResNet-based feature extractor, to ultimately predict the number of people in the scene. The target dataset (JHU-crowd) contains images with 0-10,000 people, and thus, we restricted the training samples of the source dataset (NWPU) to those that have less than 10k humans in each image. Table~\ref{table:people-datasets} shows the train-test-validation split of the datasets.

\begin{table}[!ht]
\caption{
Training, validation and test sets for each dataset. The same train-test-validation split was used, as in the original benchmark.
}
\label{table:people-datasets}
\centering
\resizebox{\linewidth}{!}{
\begin{tabular}{lccc}
\toprule 

{\textbf{Dataset}} & \textbf{Train} & \textbf{Validation} & \textbf{Test} \\ \midrule

NWPU (Source)       &  3,100  &   499   &  1,500  \\
JHU-crowd (Target)        &  2,269  &   500   &  1,600   \\
\bottomrule
\end{tabular} %
}
\end{table}

\noindent{\bf Models}. 
Figure~\ref{figure: Cancer and People Models}A shows the model used for people counting. The patches from each image were passed through a ResNet50~\cite{resnet50} model, pretrained on ImageNet~\cite{imagenet}, to get six 2048-dimensional encoding vectors. These vectors were then combined using an ``attention'' head to obtain a 64-dimensional embedding for the image. This was then fed to a dense layer to predict the number of people in the scene.

\subsection{Tumor size estimation from histopathology images}
We also addressed the challenge of estimating tumor sizes from cancer histopathology images.
\label{appendix: tumor-size prediction data}

\noindent{\bf Preprocessing}. We employed 
the Camelyon-16~\cite{cam16} and Camelyon-17~\cite{cam17} high-resolution breast-cancer datasets. The patch level annotations available in these datasets render them particularly suitable for our analysis. We used 256x256 high-resolution patches from the whole-slide images (WSIs) for our analysis. The annotations in these patches were used to compute the fraction of the patch that has tumerous tissue. This fraction was predicted as the ``tumor size'' at the patch level. Table~\ref{table:cancer-datasets} shows the train-validation-test split for the source and target datasets. To render the transfer more challenging, we used only patches containing 20-80\% tumor in the image for training the source model, while the target patches had tumor coverage ranging between 0.1-99.9\%.

\begin{table}[!ht]
\caption{
Training, validation and test sets for each dataset. A 64-16-20\% stratified split was used for both datasets.
}
\label{table:cancer-datasets}
\centering
\resizebox{\linewidth}{!}{
\begin{tabular}{lccc}
\toprule 

{\textbf{Dataset}} & \textbf{Train} & \textbf{Validation} & \textbf{Test} \\ \midrule

Camelyon-16 (Source)       &  65,337  &   16,335   &  20,418  \\
Camelyon-17 (Target)        &  7,926  &   1,982   &  2,477   \\
\bottomrule
\end{tabular} %
}
\end{table}

\begin{figure}
\begin{center}
\includegraphics[scale=0.33]{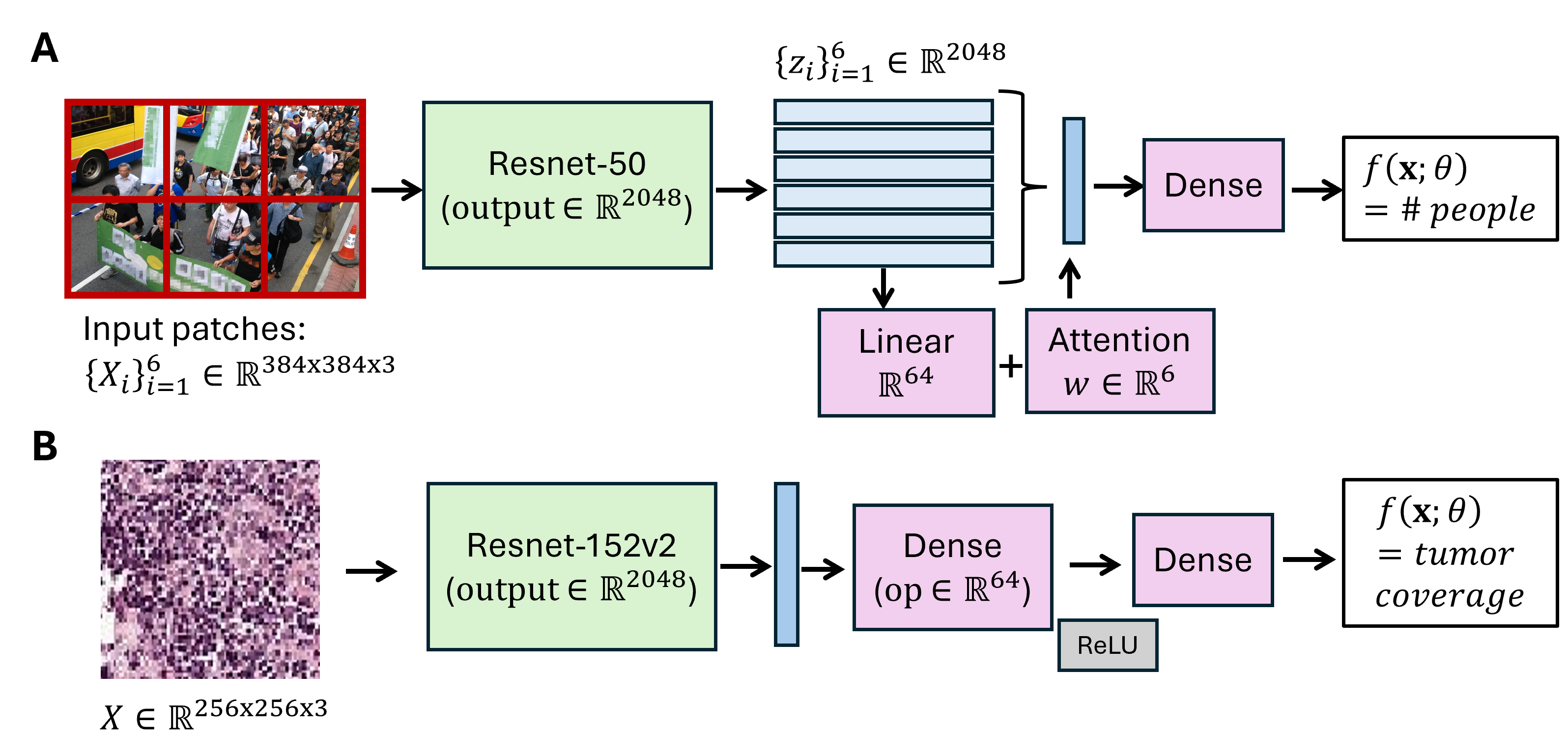}
\end{center}
\caption{Model Architectures. ({\bf A}) Predicting the number of people in a scene. The Resnet50 model is used to extract features from 6 patches of an image, along with an ``attention'' head that combines them, followed by a dense layer to make predictions; ({\bf B}) Predicting the fraction of tumor cells in a histopathology image. The Resnet152v2 model is used to extract features, followed by predictions using dense layers.}
\label{figure: Cancer and People Models}
\vspace{-0.4cm}
\end{figure}

\noindent {\bf Models}. The 256x256 patch was passed through a Resnet152v2~\cite{resnet152} model initialized to the ImageNet~\cite{imagenet} weights (Fig.~\ref{figure: Cancer and People Models}B). Next, the extracted features were passed through two dense layers to predict the fraction of tissue covered by the tumor.

\section{Additional Benchmarks}

\subsection{People counting}
\label{appendix: people counting}
We attempted transfer with models trained on a people counting task.
First, we trained the Attn-Resnet model to count people with the NWPU dataset. Source predictions were reasonably effective, with RMSE=459.01 and R=0.74. Next, this pretrained model was transferred to the JHU-crowd dataset (see Appendix~\ref{appendix: People Counting Methods}). The upper 2 rows of Table~\ref{table: people-count sf-ssda} show the performance when all the data was used for finetuning. All models were trained with mini-batches of 16 samples, and only the best-performing checkpoint on the held-out validation set was used for evaluating the test set. Adam optimizer with an initial learning rate of $10^{-4}$ was used. Based on observations in the previous experiments (Section~\ref{section: Experiments}) $\alpha=0.1$ was used as the weight for the unsupervised loss relative to the supervised loss.

To demonstrate the efficacy of CRAFT for SF-SSDA, target labels from the training data of the JHU-crowd dataset were removed randomly, but in a stratified manner. Figure~\ref{figure:semisup-all}A shows how the performance of these models is affected by reducing the fraction of labeled data ($n_{ul}/N$ is the fraction of unlabeled samples). A lower proportion of labeled data worsens the performance of all the models (RMSE increases), but CRAFT outperforms other DA algorithms. Surprisingly, naive transfer learning performed better than SOTA algorithms for this problem. On further investigation, we observed that the model trained on the NWPU dataset readily generalized to the JHU-crowd dataset (RMSE=451.62, R=0.75), already performing close to the ceiling. 

Figure~\ref{figure:semisup-all}B shows the prediction of people count in a scene for the best CRAFT-based Attn-Resnet model. Table~\ref{table: people-count sf-ssda} (lower 6 rows) summarizes the performance of CRAFT when 98\% of the data were unlabeled;  CRAFT outperforms SOTA SF-SSDA models by $>5\%$ in terms of RMSE. But supervised finetuning (TL) beat CRAFT by 17\%. The correlation coefficient  (R) also showed similar trends (Fig.~\ref{figure: All Rs}C). Several SF-SSDA methods rely on predicting pseudo-labels from unlabeled target data, and it is possible that subpar pseudo-label prediction exacerbated the poor performance of these algorithms.

\begin{table}[!h]
\caption{
Source Free Semi-supervised Deep Transfer of the model trained on the NWPU crowd dataset to the JHU dataset. The upper part of the table (rows 1-3) shows the performance ceiling when all labeled data in the target is used for training. The lower part (rows 4-9) shows SF-SSDA results when only 2\% of the target samples were labeled (3 seeds). 
}
\label{table: people-count sf-ssda}
\centering
\resizebox{1.01\linewidth}{!}{%
\begin{tabular}{lcc}
\toprule 
{\textbf{Method}} & \textbf{R} $\uparrow$ & \textbf{RMSE } (in \#people) $\downarrow$\\ \midrule
Naive Baseline                      &            -           &   724.17 $\pm$ 1.11   \\
Attn-Resnet  (TL, 100\%)  &    0.83 $\pm$ 0.02     &   429.25 $\pm$ 5.63    \\
\hline
{\it SF-SSDA} (2\% labels) \\
Attn-Resnet + TL           &    \bf{0.73 $\pm$ 0.01}        &   \bf{459.05 $\pm$ 4.41}    \\
Attn-Resnet + Mixup        &    0.34 $\pm$ 0.04       &   702.05 $\pm$ 8.26     \\
Attn-Resnet + BBCN         &    0.32 $\pm$ 0.06        &   1221.36 $\pm$ 237.66     \\
Attn-Resnet + TASFAR       &    0.27 $\pm$ 0.03        &   620.55 $\pm$ 37.28     \\
Attn-Resnet + DataFree     &    0.67 $\pm$ 0.03        &   582.53 $\pm$ 6.63    \\
Attn-Resnet + $\textcolor{red}{\text{CRAFT}}$    &  \it{0.69 $\pm$ 0.02}  &  \it{550.77 $\pm$ 9.86}   \\
\bottomrule
\end{tabular} %
}
\end{table}

\subsection{Tumor size estimation}
\label{appendix: tumor-size}

Finally, we attempted transfer with models trained to estimate the fraction of tumor tissue in a high-resolution histopathological slide. The source model (Resnet152v2) was trained to predict what fraction of the patched Camelyon-16 dataset contained a tumor. Source predictions were fairly accurate, with RMSE=0.19 and R=0.46. Next, the model was transferred to make predictions on the Camelyon-17 dataset (see Appendix~\ref{appendix: tumor-size prediction data}). All models were trained with mini-batches of 32 samples, and only the best-performing checkpoint on the held-out validation set was used for evaluating the test set. Adam optimizer with an initial learning rate of $10^{-4}$ was used. As before, $\alpha=0.01$ was used as the unsupervised loss weightage relative to the supervised loss.

\begin{figure}[!h]
\begin{center}
\includegraphics[scale=0.33]{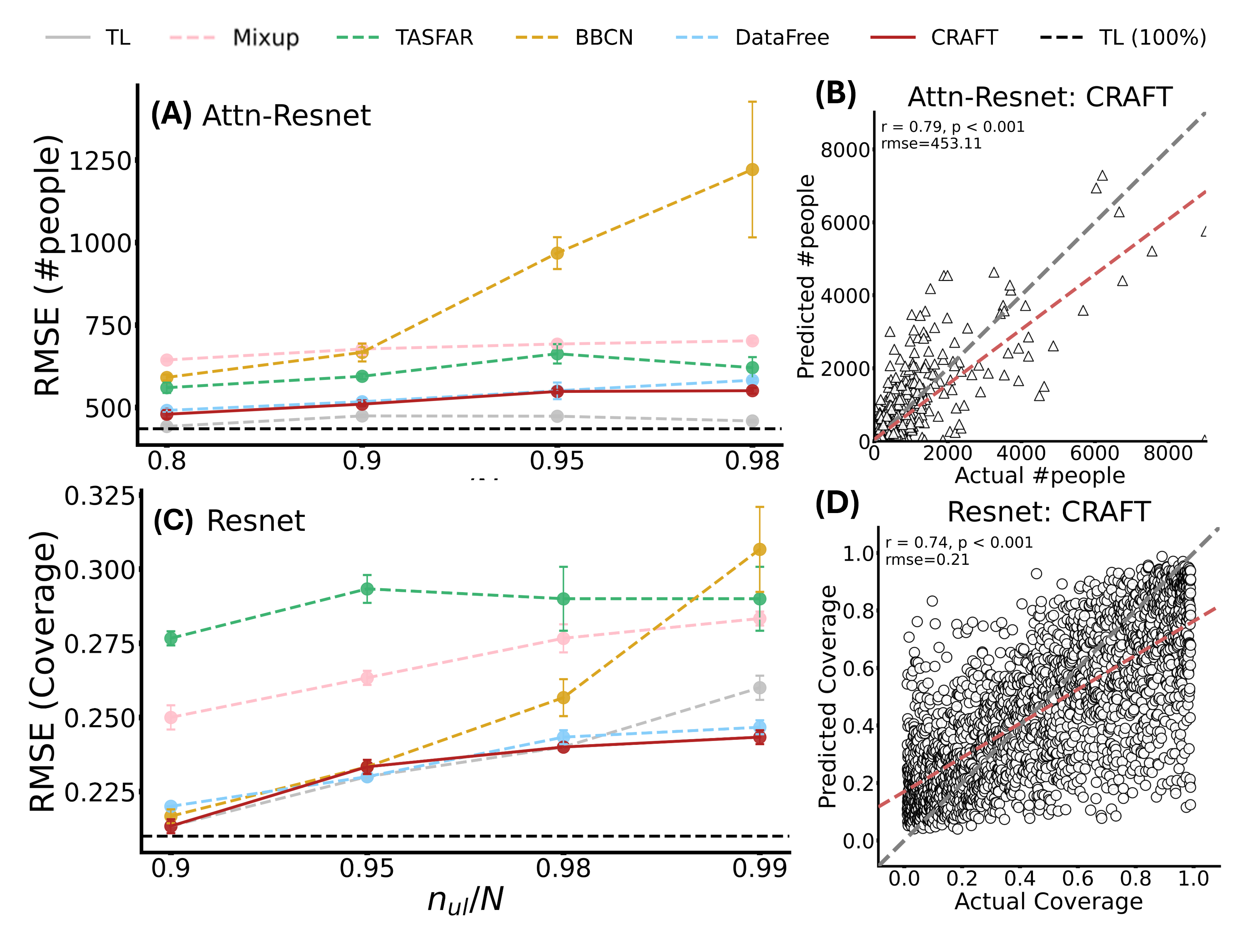}
\end{center}
\caption{({\bf A, C}) RMSE ($\downarrow$) for source-free semi-supervised domain adaptation of people counting ({\bf A}) and tumor size ({\bf C}) prediction tasks, trained with varying proportions of unlabeled target data $n_{ul}/N$. Dashed black line: Performance ceiling. ({\bf B, D}) Predictions for the best CRAFT model for people counting ({\bf B}) and tumor size prediction ({\bf D}), with 80\% and 90\% unlabeled data, respectively.
}
\label{figure:semisup-all}
\end{figure}

\begin{figure}
\begin{center}
\includegraphics[scale=0.33]{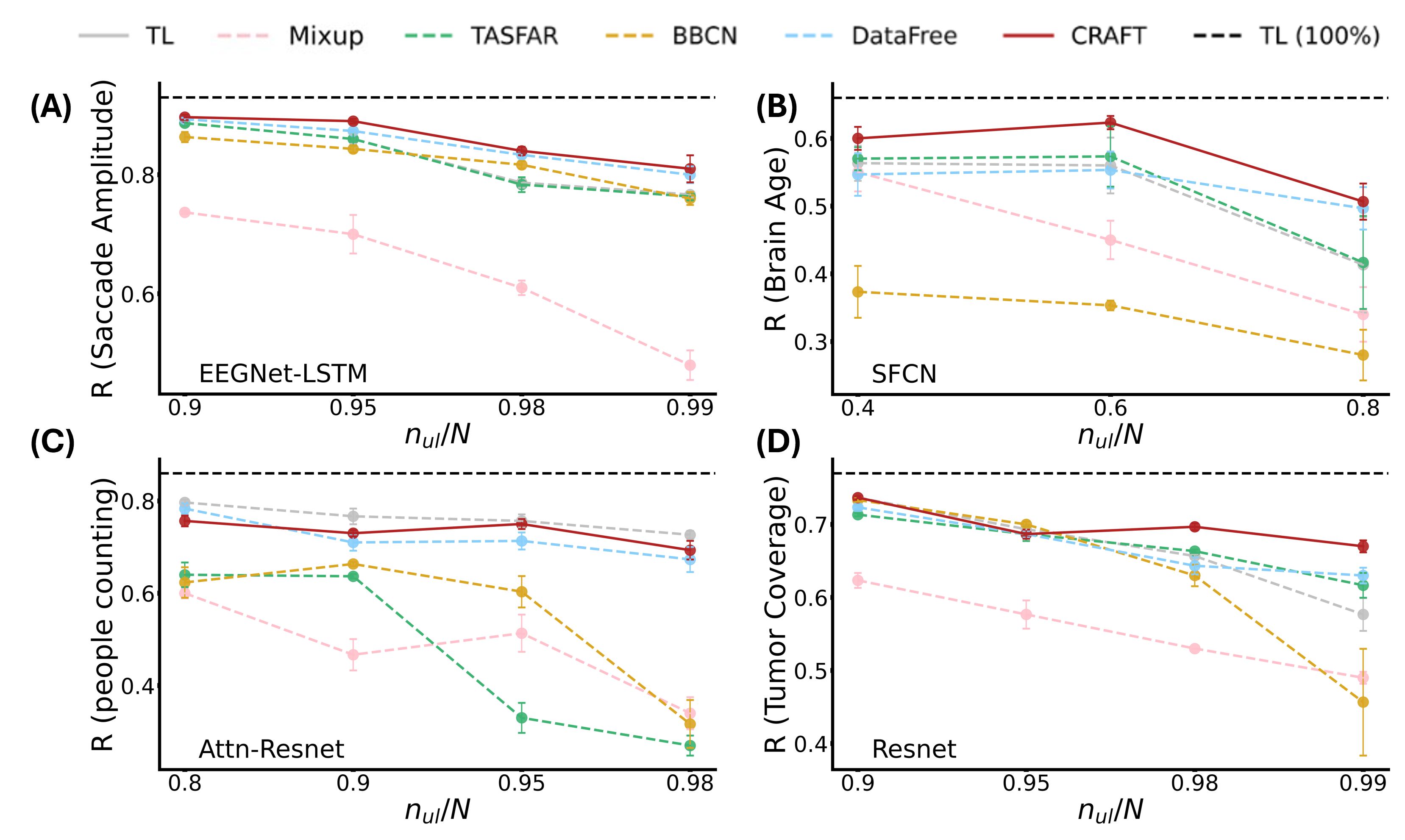}
\end{center}
\caption{Correlation Coefficient R ($\uparrow$) for all the competing SF-SSDA models at predicting ({\bf A}) Saccade Amplitude; ({\bf B}) Brain Age; ({\bf C}) Number of People; and ({\bf D}) Tumor Coverage.}
\label{figure: All Rs}
\vspace{-0.4cm}
\end{figure}

The top of Table \ref{table: tumour sf-ssda} (row 2) quantifies the performance ceiling when all the training data are used (3 seeds). In rows 3-8, we compare the performance of all the source-free models when only 1\% of the target dataset was labeled. We observe that CRAFT outperformed all SF-SSDA methods by $> 6\%$ on the correlation coefficient (R). Only DataFree performs comparably in terms of RMSE. Figure~\ref{figure:semisup-all}C shows that CRAFT performs better than all competing methods when the fraction of unlabeled data is varied from 90-99\%. Similar trends were observed for the correlation coefficient R (Fig.~\ref{figure: All Rs}D). Figure~\ref{figure:semisup-all}D shows the prediction scatter of the best-performing CRAFT model when 90\% of the data was labeled.

\begin{table}[!h]
\caption{
Source Free Semi-supervised Deep Transfer of the model trained on Camelyon-16 dataset to the Camelyon-17 dataset. The upper part of the table (rows 1-3) shows the performance ceiling when all labeled data in the target is used for training. The lower part (rows 4-9) shows SF-SSDA results when only 1\% of the target samples were labeled (3 seeds). 
}
\label{table: tumour sf-ssda}
\centering
\resizebox{1.01\linewidth}{!}{%
\begin{tabular}{lcc}
\toprule 
{\textbf{Method}} & \textbf{R} $\uparrow$ & \textbf{RMSE } $\downarrow$\\ \midrule
Naive Baseline        &            -                &   0.313 $\pm$ 0.003   \\
Resnet (TL, 100\%)    &    0.773 $\pm$ 0.003        &   0.203 $\pm$ 0.003   \\
\hline
{\it SF-SSDA} (1\% labels) \\
Resnet + TL           &    0.577 $\pm$ 0.026        &   0.260 $\pm$ 0.005    \\
Resnet + Mixup        &    0.490 $\pm$ 0.009        &   0.283 $\pm$ 0.003     \\
Resnet + BBCN         &    0.457 $\pm$ 0.084        &   0.307 $\pm$ 0.017     \\
Resnet + TASFAR       &    0.617 $\pm$ 0.020        &   0.290 $\pm$ 0.012    \\
Resnet + DataFree     &    \it{0.630 $\pm$ 0.012}        &   \it{0.247 $\pm$ 0.003}    \\
Resnet + $\textcolor{red}{\text{CRAFT}}$    &  \bf{0.670 $\pm$ 0.009}  &  \bf{0.243 $\pm$ 0.003}   \\
\bottomrule
\end{tabular} %
}
\end{table}

\section{Analysis of hyperparameters: bin size, $\alpha$}
\label{Appendix: More Hyperparameter details}
\noindent{\bf Bin Size.} We varied the number of bins in the pseudo-label selection step of CRAFT to quantify the effect of bin sizes. We found the algorithm to be fairly robust to the choice of bin sizes for both the neuroscience datasets (Table~\ref{table:Binsize}).

\begin{table}[h!]
\caption{
 Effect of bin size on the performance of CRAFT for SF-SSDA for the saccade prediction (5\% labeled) and brain-age prediction (60\% labeled) tasks.}
\label{table:Binsize}
\centering
\resizebox{1.0\linewidth}{!}{%
\begin{tabular}{lcccc}
\toprule 
 & \multicolumn{2}{c}{\it{\textcolor{gray}{\bf EEGNet-LSTM} (P $\sim$ 190K)}}  & \multicolumn{2}{c}{\it{\textcolor{gray}{\bf SFCN} (P $\sim$ 2.95M)}} \\ 
{\textbf{Bin Size}} & \textbf{R} $\uparrow$ & \textbf{RMSE} $\downarrow$  &  \textbf{R} $\uparrow$  & \textbf{RMSE} $\downarrow$  \\
$\%$ Range(y) &  &   &   &  \\
\midrule
0.50                                    &   0.88 $\pm$ 0.01    &   67.25 $\pm$ 0.46    &   0.65 $\pm$ 0.03    &    6.53 $\pm$ 0.15   \\
0.33                                    &   0.89 $\pm$ 0.01    &   65.84 $\pm$ 0.98    &   0.55 $\pm$ 0.02    &    6.91 $\pm$ 0.09   \\
0.25                                    &   0.89 $\pm$ 0.01    &   67.39 $\pm$ 1.39    &   0.62 $\pm$ 0.01    &    6.60 $\pm$ 0.02   \\
0.20                                    &   0.89 $\pm$ 0.01    &   65.21 $\pm$ 0.89    &   0.64 $\pm$ 0.02    &    6.36 $\pm$ 0.09   \\
\bottomrule

\end{tabular} %
}
\end{table}

\noindent{\bf Unsupervised loss weight ($\alpha$).} For the Saccade prediction task, a cross-validation set was available and hence, only the best-performing model, based on the corresponding $\alpha$, is reported. However, a validation set was unavailable for brain age prediction. Hence, we report the performance across three $\alpha$ values $\{0.01, 0.1, 1.0\}$ in Table~\ref{table:Brain-Age Predictions: alphas}. While CRAFT outperforms competing models for low values of alpha, large weightages to the unsupervised loss (e.g., $\alpha=1$) result in comparatively poor performance.

\begin{table}[h!]
\caption{
Effect of $\alpha$ (unsupervised loss weight) on the performance of CRAFT and other competing methods for SF-SSDA for the brain-age prediction (20\% labeled) tasks.}
\label{table:Brain-Age Predictions: alphas}
\centering
\resizebox{1.0\linewidth}{!}{%
\begin{tabular}{lcccccc}
\toprule 
 & \multicolumn{2}{c}{TASFAR}  & \multicolumn{2}{c}{DataFree} & \multicolumn{2}{c}{\textcolor{red}{\bf CRAFT}}  \\ 
{\textbf{$\alpha$}} & \textbf{R} $\uparrow$ & \textbf{RMSE} $\downarrow$  &  \textbf{R} $\uparrow$  & \textbf{RMSE} $\downarrow$ &  \textbf{R} $\uparrow$  & \textbf{RMSE} $\downarrow$  \\
\midrule
0.01  &   0.40 $\pm$ 0.07    &   7.53 $\pm$ 0.18    &   0.31 $\pm$ 0.04   &   7.82 $\pm$ 0.20 &  \bf{0.51 $\pm$ 0.03} & \bf{7.31 $\pm$ 0.17}  \\
0.10  &    0.42 $\pm$ 0.07    &   7.47 $\pm$ 0.15    &   0.50 $\pm$ 0.05   &    7.35 $\pm$ 0.15  &  \bf{0.51 $\pm$ 0.03}  &   \bf{7.14 $\pm$ 0.11}  \\
1.00  &    \bf{0.40 $\pm$ 0.06}    &   \bf{7.47 $\pm$ 0.21}    &   0.07 $\pm$ 0.02    &    8.70 $\pm$ 0.06  &  0.37 $\pm$ 0.03  & 8.64 $\pm$ 0.15 \\
\bottomrule

\end{tabular} %
}
\end{table}

\section{Further Details}
\subsection{Compute Resources and Software}
\label{appendix: Resources and Software}
All the algorithms were implemented using TensorFlow 2.x, and experiments were run on 24 GB RTX-4090Ti, 32 GB V100, and 40 GB A100 GPUs.

\subsection{Code Availability}
\label{appendix: code}
The implementation of all the methods described in the study is publicly available in the following repository -- \url{https://github.com/mainak-biswas1999/CRAFT_ICCV2025.git}.

\end{document}